\documentclass[letterpaper, 10 pt, conference]{ieeeconf}  

\IEEEoverridecommandlockouts                         
\usepackage{todonotes}
\usepackage{amsmath}
\usepackage{amssymb}
\usepackage{bbm}
\usepackage{tabularx}
\usepackage{dsfont}

\usepackage{hyperref}
\usepackage{url}
\usepackage{graphicx}
\usepackage{subcaption}

\usepackage{caption}
\usepackage{amssymb}

\usepackage{amsmath}
\usepackage{amsfonts}
\usepackage[english]{babel}
\usepackage[bottom]{footmisc}
\usepackage{wrapfig}
\usepackage[ruled]{algorithm}
\usepackage{algpseudocode}
\usepackage{amsthm}
\usepackage{siunitx}
\usepackage{array}
\usepackage{color}
\usepackage{ifthen}

\newcommand{\extratext}[1]{} 
\newcommand{\dataset}[0]{ \mathcal{D} }
\newcommand{\wv}[0]{ \mathbf{w} }
\newcommand{\numclasses}[0]{ C }

\newtheorem{definition}{Definition}

\newcommand{\LL}[1]{{\color{red}[LL: #1]}}
\newcommand{\MW}[1]{{\color{blue}[MW: #1]}}
\newcommand{\MK}[1]{{\color{magenta}[MK: #1]}}
\newcommand{\RM}[1]{{\color{cyan}[RM: #1]}}

\newboolean{showoldcomments} 
\setboolean{showoldcomments}{false} 

\title{\LARGE \bf
Uncertainty Quantification with Statistical Guarantees in End-to-End Autonomous Driving Control
}

\author{Rhiannon Michelmore$^{1}$, Matthew Wicker$^{1}$, Luca Laurenti$^{1}$, Luca Cardelli$^{1}$, Yarin Gal$^{1}$, Marta Kwiatkowska$^{1}$  
\thanks{*This work was not supported by any organization}
\thanks{$^{1}$Department of Computer Science, University of Oxford, United Kingdom. {\tt\small firstname.lastname@cs.ox.ac.uk}}
}

\begin{document}

\maketitle
\thispagestyle{empty}
\pagestyle{empty}


\begin{abstract}

Deep neural network controllers for autonomous driving have recently benefited from significant performance improvements, and have begun deployment in the real world. Prior to their widespread adoption, safety guarantees are needed on the controller behaviour that properly take account of the uncertainty within the model as well as sensor noise. Bayesian neural networks, which assume a prior over the weights, have been shown capable of producing such uncertainty measures, but properties surrounding their safety have not yet been quantified for use in autonomous driving scenarios. In this paper, we develop a framework based on a state-of-the-art simulator for evaluating end-to-end Bayesian controllers. In addition to computing pointwise uncertainty measures that can be computed in real time and with statistical guarantees, we also provide a method for estimating the probability that, given a scenario, the controller keeps the car safe within a finite horizon. We experimentally evaluate the quality of uncertainty computation by 
three Bayesian inference methods in different scenarios and show how the 
uncertainty measures can be combined and calibrated for use in collision avoidance. Our results suggest that uncertainty estimates can greatly aid decision making in autonomous driving.  

\end{abstract}

\section*{Introduction}

Deep Neural Networks (DNNs) have seen a surge in popularity over the past decade, and their use has become widespread in many fields 
including safety-critical systems such as medical diagnosis and, in particular, autonomous cars. The latter have driven millions of miles without human intervention \cite{Waymo,CADis}, 
but offer few safety guarantees. 
This has led to erroneous edge-case behaviours and unforeseen consequences \cite{TeslaCrash}. 
Thus, there is an urgent need for methods that are capable of accurately detecting, analysing and diagnosing such erroneous behaviours.

A Bayesian Neural Network (BNN) is a neural network with a prior distribution on its weights. BNNs have the  ability to capture the uncertainty within
the learning model, while retaining the main advantages intrinsic to deep neural networks \cite{mackay1992practical}. 
As a consequence, they are particularly appealing for safety-critical applications, such as autonomous driving, where
uncertainty estimates can be propagated through the decision
pipeline to enable safe decision making \cite{mcallister2017concrete}.
Consider, for example, a self-driving car that, while driving, finds an obstacle in the middle of the road. Then, the controller may be uncertain on the steering angle to apply and, in order to avoid the obstacle, may choose angles which turn the car either right or left, with equal probability.
Nevertheless, if we consider the optimal decision according to this steering angle distribution and a squared loss, then the controller will simply select the mean value of the distribution \cite{bishop2006pattern} and aim straight at the obstacle. As we will show later (Definition \ref{def:ProblemRealTIme}), having precise quantitative measures of the BNN uncertainty
facilitates the detection of 
such ambiguous situations.

In this paper we develop a novel framework for evaluating the safety of autonomous driving using end-to-end BNN controllers, that is, 
controllers in which the end-to-end process, from sensors to actuation, involves a single BNN without modularisation. 
Our framework can be configured with any simulator and assumes that trajectories can be sampled efficiently and are endowed with a probability measure. 
We 
demonstrate how to obtain \emph{a priori} statistical guarantees on the safety of the application of the BNN in a given scenario. In particular,  we consider both \emph{probabilistic safety}, which is the probability that the controller will keep the car safe for a given time horizon, and \emph{real-time decision confidence}, which is the probability that the BNN  is certain of a given decision. By using concentration inequalities, such as Chernoff bounds \cite{chernoff1952measure}, we show that both measures can be estimated with arbitrarily stringent \emph{a priori} guarantees. 

We evaluate our methods on experiments performed on the CARLA driving simulator \cite{Dosovitskiy17}, where we consider a deep end-to-end controller given by a modified  NVIDIA's PilotNet (formally  known  as  DAVE-2)  neural network architecture \cite{bojarski2016end}, which we train with three different BNN inference methods, Monte Carlo
dropout \cite{gal2016dropout}, mean-field variational inference \cite{blundell2015weight}, and Hamiltonian Monte Carlo \cite{neal2011hmc}.
We consider different training scenarios, including obstacle avoidance and driving on a roundabout, 
demonstrating how to quantify the uncertainty of the controller's decisions and utilise uncertainty thresholds in order to guarantee the safety of the self-driving car with high probability.  
In summary, this paper makes the following main contributions:
\begin{itemize}
    \item We present a framework for evaluating safety of autonomous driving with end-to-end BNN controllers, which is based on a simulator and allows one to obtain and quantify the quality of uncertainty estimates for the controller's decisions. 
    \item We design a statistical framework for evaluating safety of BNN controllers with high probability with a priori statistical guarantees.
    \item We show that this statistical framework can be used to evaluate model robustness to changes in weather, location, and observation noise. 
    \item We empirically demonstrate that our real-time statistical estimates can be used to avoid a high percentage of collisions. 
\end{itemize}

\section{Related Works}

Deep end-to-end controllers are rising in popularity as the state-of-the-art method for autonomous driving. Examples of such controllers include CNNs,  \cite{chen2017end} and \cite{bojarski2017explaining}, and fully convolutional networks with long short term memory (FCN-LSTM), \cite{xu2017end}.
Prior to end-to-end controllers, there is a rich literature on detecting anomalies from sensor output \cite{isermann1984faultdetection}; however, these methods deal with when sensor outputs deviate from normal ranges and do not detect when the model itself is unsafe. For this, quantification of model and data uncertainty, extracted from BNNs, can be used \cite{kendall2017uncertainties}. 

To date, the advantages of BNNs have been observed in small test cases. In \cite{lee2018ensemble}, an ensemble of BNNs over different modalities (stereo imaging and GPS) are used in order to drive a 1:5 scale car around an oval track. Further, in \cite{kahn2017uncertainty}, they use bootstrapping and dropout in order to generate uncertainty estimates which allow an RC car or quad-rotor drone to predict and avoid collisions. 

Beyond these simplified domains important work is being done in scaling end-to-end BNN models to real-world test cases. In \cite{amini2019variational}, the authors use a BNN to incorporate GPS and image data to make predictions about long term navigation and localization. \cite{huang2019uncertainty} looks at using uncertainty from a BNN to produce both a distribution of possible future trajectories of a car at an intersection, and a confidence estimate for varying time horizons, with the final goal of augmenting the result of this with a physics-based predictor using confidence estimates. Additionally, in \cite{feng2018towards} BNNs are used on real-world LiDAR data in order to more safely localize objects.

While these works do well to scale BNNs to more pratical cases, they are not concerned with analysis of the safety of deployment for BNNs. For this, very few works have been completed. \cite{quilbeuf2018statistical} looks at using statistical model checking (SMC) to evaluate the probability of two different subsystems of an autonomous vehicle controller (therefore not an end-to-end controller) meeting specific key performance indicators (KPIs). Although the results of this paper demonstrated a high probability of meeting the KPIs, the simulator used lacked realistic detail.

We further the investigation into safe deployment of BNNs as end-to-end controllers by scaling exact and approximate inference techniques to realistic simulators. This allows for the contextualization of pointwise uncertainty estimates and enables their use in real-time decision making. Understanding that uncertainty increases for certain inputs (as in \cite{amini2019variational, feng2018towards,huang2019uncertainty}) is important insofar as it encourages the use of uncertainty during deployment; however, evaluating the uncertainty in a pointwise (per image) fashion does not allow us to reason about emergent properties of the incorporation of uncertainty and their safety \cite{cardelli2019robustness}. In order to create safe plans for autonomous vehicles that incorporate uncertainty, we must evaluate the fundamental impact of decisions which are made on the basis of uncertainty (e.g. slowing down when uncertain, or returning control to the user).

\section{Background}

\ifthenelse{\boolean{showoldcomments}}{\MK{POMDPs come out of the blue and their role is not clear (here and in the rest of the paper). This needs to be strengthened, and the relationship of BNNs and POMDPs addressed.}\LL{Agreed. And I believe we do not even need them.}
\LL{I am trying to re-write everything without POMDPs. I fear that if we introduce POMDPs, the reviewers may want to see that our strategy minimize some kind of reward or they may expect we use some of their properties. Also, we do not really use them. We just need to assume there exists a well defined probability measure. So, I think we can be more general and avoid unwanted comments.}\LL{In the future, we should consider POMDPs and show that approximated strategies with neural networks can be built}}{}

\subsection{Bayesian Neural Networks and Inference}\label{Sec:BNN}

For a test input $o\in \mathbb{R}^m$ a BNN  with $\numclasses$ output units and an unspecified number (and kind) of hidden layers is denoted as ${f}^\wv(o) = [f_1^\wv(o),\ldots,f_\numclasses^\wv(o)]$, where $\wv$ is the weight vector random variable.
Given  $w $, a weight sampled from the distribution of $\wv$, we denote with ${f}^w(o)$ the corresponding deterministic neural network with weights fixed to $w$ and with $p(f^{\mathbf{w}}(o))$ the resulting distribution of ${f}^{\mathbf{w}}(o)$.
 In the case of classification, we consider classification with a softmax likelihood model. 
Let $ \dataset =\{(o,c) \, | \, o \in \mathbb{R}^m,\; c \in \{1,...,\numclasses \} \}$ be the training set. Then, we assume a prior distribution over the weights, i.e.\ $ \wv \sim p(w)$\footnote{Usually depending on hyperparameters,  omitted here for simplicity.}, so that learning for the BNN amounts to computing the posterior distribution over the weights, $p (w \vert \dataset )$, via the application of Bayes rule. 
Unfortunately, because of the non-linearity introduced by the neural network architecture, the computation of the posterior cannot be done analytically \cite{mackay1992practical}.
Hence, various approximation methods have been studied to perform inference with BNNs in practice.
Among these methods, we consider 
Hamiltonian Monte Carlo (HMC) \cite{neal2011hmc}, Mean Field Variational Inference (VI) \cite{blundell2015weight} \cite{graves2011practical}, and Monte Carlo Dropout (MCD) \cite{gal2016dropout}.

\textbf{Hamiltonian Monte Carlo (HMC)} 
proceeds by defining a Markov chain whose invariant distribution is $p(w \vert \dataset)$, and relies on Hamiltionian dynamics to speed up the exploration of the space.
Differently from the two other methods discussed below, HMC does not make any assumptions on the form of the posterior distribution, and is asymptotically correct.
The result of HMC is a set of samples $w_i$ that approximates  $p(w \vert \dataset)$.

\textbf{Mean Field Variational Inference (VI)} 
proceeds by finding a Gaussian approximating distribution $q(w) \approx p(w \vert \dataset)$ in a trade-off between approximation accuracy and scalability.
The core idea is that $q(w)$ depends on some hyper-parameters that are then
iteratively optimized by minimizing a divergence measure between $q (w)$ and $ p (w | \dataset)$. 
Samples can then be efficiently extracted from $q(w)$.

\textbf{Monte Carlo Dropout (MCD)} 
is an approximate variational inference method based on dropout \cite{gal2016dropout}.  
The approximating distribution $q(w)$ takes the form of the product between Bernoulli random variables and the corresponding weights. 
Hence, sampling from $q(w)$ reduces to sampling Bernoulli variables, and is thus very efficient.

\section{Uncertainty Quantification for Autonomous Driving}
In this section we first give a 
description of our 
framework for evaluating BNN controllers and then 
introduce different measures for safety characterization in self-driving cars. In particular, in Definition \ref{Def:PlanningTImeHorizon} we define \emph{probabilistic safety}, which is the probability that a BNN controller will keep the car safe, while in Definition \ref{def:ProblemRealTIme} we define \emph{real-time decision confidence} as the probability that the BNN controller is certain of its decision at the current time.

\subsection{Conceptual Description of our Framework}
We model the autonomous driving scenario considered in this paper as a discrete-time controlled stochastic process ($\mathbf{x}_k, k \in \mathbb{N})$ \cite{gihman2012controlled}.  $\mathbf{x}_k$ is a probabilistic model that describes the status of the entire system and takes values in a state space $\mathcal{X}$, which includes information on the position, velocity and acceleration of the car, as well as that of all the other vehicles, pedestrians and obstacles on the map. Intuitively, in this paper, $\mathbf{x}_k$ just represents a white-box system that we assume we can simulate arbitrarily many times.

The control space of the process, which represents the set of variables  a controller can modify to drive the behaviour of $\mathbf{x}_k$, is denoted by $\mathcal{U}\subseteq \mathbb{R}^m$ and is
typically given by steering angle, braking and acceleration values of the ego car.
We assume the controller can only observe a noisy image of the state space coming from the available sensors. 
Hence, $\mathbf{x}_k$ is only \emph{partially observable}. We denote by  $\mathcal{O}$ the observation space, which is the set of all possible observations. Intuitively, given the current state of $\mathbf{x}_k$, the controller receives an observation of $\mathbf{x}_k$, $o\in \mathcal{O},$ and  synthesizes an action $u \in \mathcal{U}$ based on this observation. Then, $\mathbf{x}_k$ transitions to a new state at time $k+1$.  Given $a$ the evolution of $\mathbf{x}_{k}$ is probabilistic, as traffic,  weather conditions, and other variables are uncertain.


A (memoryless and deterministic) control strategy for $\mathbf{x}_k$, $\pi:\mathcal{O}\to \mathcal{U}$ associates to a given observation an action. In this work, as explained in detail in the next section, we train a BNN controller to synthesize $\pi$.
 We denote a \emph{path} of $\mathbf{x}_k$ by
$\omega:\mathbb{N}\to \mathcal{X}\times \mathcal{U}$. $\omega$ is a sequence of states and actions in an execution of the system.
Given a strategy $\pi$ we assume there exists a well defined  probability measure  over the paths of $\mathbf{x}$  such that, for $X \subseteq \mathcal{X}$, 
$P( \omega(k) \in X | \pi )$
is the probability that $\mathbf{x}_k$ is in $X$ at time $k$ given $\pi$.
For instance, this measure is well defined for POMDPs
\cite{chatterjee2016decidable}. However, the uncertainty quantification techniques derived in this paper will work also for more general, possibly non-Markov, processes.

\subsection{Safety Measures for Autonoumous Driving}
The first problem we consider in Definition \ref{Def:PlanningTImeHorizon} is that of computing  the probability that a given strategy $\pi$ synthesized by the BNN keeps the car safe.
This probability can  be used for planning and to certify that a given controller is safe with high probability given the available information.
Computing this value can be done in any simulator. Prior to the deployment of an autonomous vehicle it is common for large companies to evaluate the safety of specific test cases \cite{reynolds2018uber}.
As a consequence, we believe that a quantifiable notion on the safety of a given controller is pivotal in order to certify a controller, especially if this incorporates learning elements. 


\begin{definition}\label{Def:PlanningTImeHorizon}{(Probabilistic Safety)}
Let $X\subseteq \mathcal{X}$ be a safe set, $\omega$ denote a path of $\mathbf{x}_k,$ $[0,T]\subseteq \mathbb{N}$ be a time horizon, and $\pi$ be a given policy.
Compute
\begin{align*}
&    \eta_1=P(\phi_1(\pi,\omega,[0,T])), \\
&\text{ where }  \phi_1(\pi,\omega,[0,T])=\forall k\in [0,T], \omega(k) \in X\, | \, \pi
\end{align*}
Then, for $\delta>0$, we say that  $\pi$ is $\delta-$safe in $[0,T]$ iff
$\eta_1\geq \delta.$
\end{definition}

\noindent
$\eta_1$ is satisfied if the probability that a path of $\mathbf{x}_k$ is safe during the interval $[0,T]$ is greater than a threshold. 
We should also stress that similar probabilistic measures of safety are widely used to certify  cyber-physical system models \cite{abate2008probabilistic,Bortolussi:2019:CLM:3347091.3331452}. 

 As explained in greater detail in the next section, in order to synthesize a control strategy $\pi$, we train a BNN and we obtain that, for an image $o\in \mathcal{O}$, $\pi(o)$ is determined by the BNN predictions.
 However, notice that $\pi(o)$ is still deterministic. Hence, it does not take into account the uncertainty in the model predictions, which is intrinsic in the BNN and could be used to quantify the confidence of the model in its decisions. 
To tackle this issue, for $o\in \mathcal{O}$, in the following definition, we consider a notion of trust  of $\pi(o)$ based on the probability mass of the BNN around $\pi(o)$. The following problem is stated for regression tasks, but can be trivially extended to classification problems.

\begin{definition}\label{def:ProblemRealTIme}{(Real-time decision confidence)}
Given $\epsilon >0$ let ${o}_k$ the observation received at time $k$, $w$ a wieght sampled from $\mathbf{w},$ and  $S^{o_k}=\{u \in \mathcal{U} \,s.t.\, |u-\pi(o_k)|\leq \epsilon \}$.
Compute
\begin{align*}
    &\eta_2=P\big(\phi_2(S^{o_k},f^{{w}}(o_k)) \big), \\
    &\text{where }\phi_2(S^{o_k},f^{{w}}(o_k))=f^{{w}}(o_k) \in S^{o _k}.
\end{align*}
Then, we say that the decision at time $k$ is $\delta-$confident iff
$ \eta_2\geq \delta. $
\end{definition}

Note that the probability measure in the above definition comes from the distribution of the weights in the BNN. In fact, by definition of probability, we can equivalently write $\eta_2=\mathbb{E}_{w\sim \mathbf{w}}[ \mathbf{1}_{f^{w}(o_k) \in S^{o _k}}] $, where $\mathbf{1}_E$ is the indicator function for event $E$.
Hence, real-time decision confidence, as defined in Definition \ref{def:ProblemRealTIme}, seeks to compute the probability mass in a $\epsilon-$ball around $\pi(o)$ and classify a decision as certain if the resulting probability is greater than a threshold.
 Definition \ref{def:ProblemRealTIme} can be violated either when there is high uncertainty (i.e., variance is large) or when the control distribution is multimodal and the most likely mode of $p(f^{\mathbf{w}})(o)$ is far from $\pi(o)$.
In the experimental results section we show that this measure of uncertainty can be employed together with commonly employed measures of uncertainty, such as \emph{mutual information} \cite{shannon2001mathematical}, to quantify in real time the degree that the model is confident in its predictions and can offer a notion of trust in its predictions.

In the next subsection we consider a statistical framework that allows us to compute the measures of  Definition \ref{def:ProblemRealTIme} and \ref{Def:PlanningTImeHorizon} with guarantees in terms of confidence intervals.



\subsection{A Statistical Framework for Safety Evaluation}\label{Subsec:stat_frame_robust}

For the computation of $\eta_1$ and $\eta_2$, we consider a statistical framework, inspired by the techniques developed for statistical analysis of probabilistic models \cite{cardelli2019statistical,legay2010statistical}. 
In particular, we observe that the satisfaction of both $\phi_1$ and $\phi_2$ can be seen as \emph{Bernoulli random variables}, which we can observe by sampling from $\mathbf{w}$, the weights of the BNN in case of real-time decision confidence, and by sampling $\mathbf{x}_k$ in case of probabilistic safety. After we collect $n$ samples of each random variable, we can build the following empirical estimators
\begin{align*}
   &\hat \eta_1=\frac{1}{N}\sum_{i=1}^n \phi_1(\pi,\omega_i,[0,T]) \\
    &\hat \eta_2=\frac{1}{N}\sum_{i=1}^n \phi_2(\pi(o),S^{o},f^{{w}_i}(o)) ,
\end{align*}
where $\{w_1,...,w_n \}$ are weights sampled from $\mathbf{w}$ and $\{\omega_1,...,\omega_n \}$ are paths sampled from $\mathbf{x}.$
Then, for an arbitrary absolute error bound $0 < \theta < 1$ 
and confidence  $0< \gamma \leq 1$, we obtain that if 
\begin{equation}\label{eq:chernoff}
n > \frac{1}{2\theta^2}\log\left(\frac{2}{\gamma}\right),
\end{equation}
then for $i\in \{1,2 \}$, it holds that
\begin{equation}\label{eq:est_constraints}
P(|\hat{\eta}_{i}-\eta_i|>\theta)\leq \gamma.
\end{equation}
The above bound is based on 
Chernoff bounds~\cite{chernoff1952measure}. Nevertheless,  also other sequential schemes, potentially requiring less samples, could be employed \cite{cardelli2019statistical}. However, the bound in Eqn \eqref{eq:chernoff} has the advantage to allow one to determine the required sample size $n$ for a given precision before performing the experiments. Hence, it can be trivially parallelized.

\section*{Bayesian End-to-End Controllers for Self Driving}\label{Sec:ControllersSelfDriving}


In the experiments considered in this paper we consider a setting where the observation space $\mathcal{O}$ is given by images from a single camera input, placed on the front centre of the car facing forwards. The control space $\mathcal{U}$ is the steering angle. Nevertheless, we should stress that the techniques developed in this paper are general and not limited to this scenario.

\subsection{Data Acquisition and Processing}
\ifthenelse{\boolean{showoldcomments}}{\RM{Do we want a nice example image from simulator here?}}{}
The experiments in this paper use the CARLA simulator, a state-of-the-art, open-source simulator for autonomous driving research \cite{Dosovitskiy17}. However, we stress that any simulator can be used within this framework, assuming it can simulate car trajectories, and generate images that can be used by the controller.  All training data, which consists of (\textit{image, steering angle}) pairs, was acquired within the CARLA simulator, either through manual driving or use of the built-in autopilot. During experiments, we also make use of the cars trajectory data, which is provided in the form of a list of GPS coordinates from the simulator. Images are converted to grayscale and scaled to a size of 64 $\times$ 48 pixels, and steering angles (recorded between -1 and 1) are binned into intervals of tenths. The data recorded consists of three scenarios: a right turn on a roundabout and a straight segment of road with and without an obstacle (stationary vehicle). It is possible to vary the weather within the simulator, however the weather condition in all of the training data is ``clear noon''.


We use a modified PilotNet \cite{bojarski2017explaining} architecture for the experiments in this paper. Traditionally, steering angle prediction has been treated as a regression problem. However, it has been shown that posing regression tasks as classification tasks often shows improvement over direct regression training \cite{rothe2015dex}. 
Therefore, we have modified the final layer of the PilotNet architecture to have neurons equal to the number of classes (variable per experiment), and a softmax activation function.

We fix the convolutional layers and first fully connected layer, and use the final layers for uncertainty extraction (similarly to \cite{ovadia2019can}). For MCD, we use concrete dropout \cite{gal2017concrete} on the final three layers (and leave the fourth fully connected). For VI and HMC, we use four fully connected layers, where the input to the first layer are the features extracted from the final fixed network layer.

In our experiments, for an observation $o$ we have that $\pi(o),$ the BNN decision, is given by the most likely class. However, we stress that other choices for $\pi(o)$ are possible according to the particular loss function (see e.g.,  \cite{bishop2006pattern}) and the methods presented in this paper are independent of the criteria for assigning $\pi(o)$.

\subsection{Network Training}

This section describes how the networks for each inference technique were trained. Full details of hyper-parameters can be found in the code associated with this work.


\textbf{MCD} The cross-entropy loss function is used, along with the ADAM optimizer with a learning rate of $0.0001$ and the dropout probabilities tuned with concrete dropout, which converged to ($0.1$, $0.08$, $0.08$). The batch size is $16$ and it was trained for $25$ epochs.

\textbf{VI} Features are first extracted from the final fixed layer of the network using the weights from the MCD network for these initial layers. Then, we impose prior distributions on the weights of the final four, fully-connected layers. These are normal distributions with mean $0$ and variable variance.
Inference was then performed 
using the Edward python library \cite{tran2016edward}, and the posterior is also in the form of a normal distribution.

\textbf{HMC} The prior distributions for the HMC networks are as above, however the posterior here is an empirical distribution based on sampling with the HMC algorithm. We use 10 steps of numerical integration prior to judging the acceptance criteria of each sample.

\section{Experiments}

\begin{figure*}
    \centering
    \includegraphics[width=0.73\textwidth]{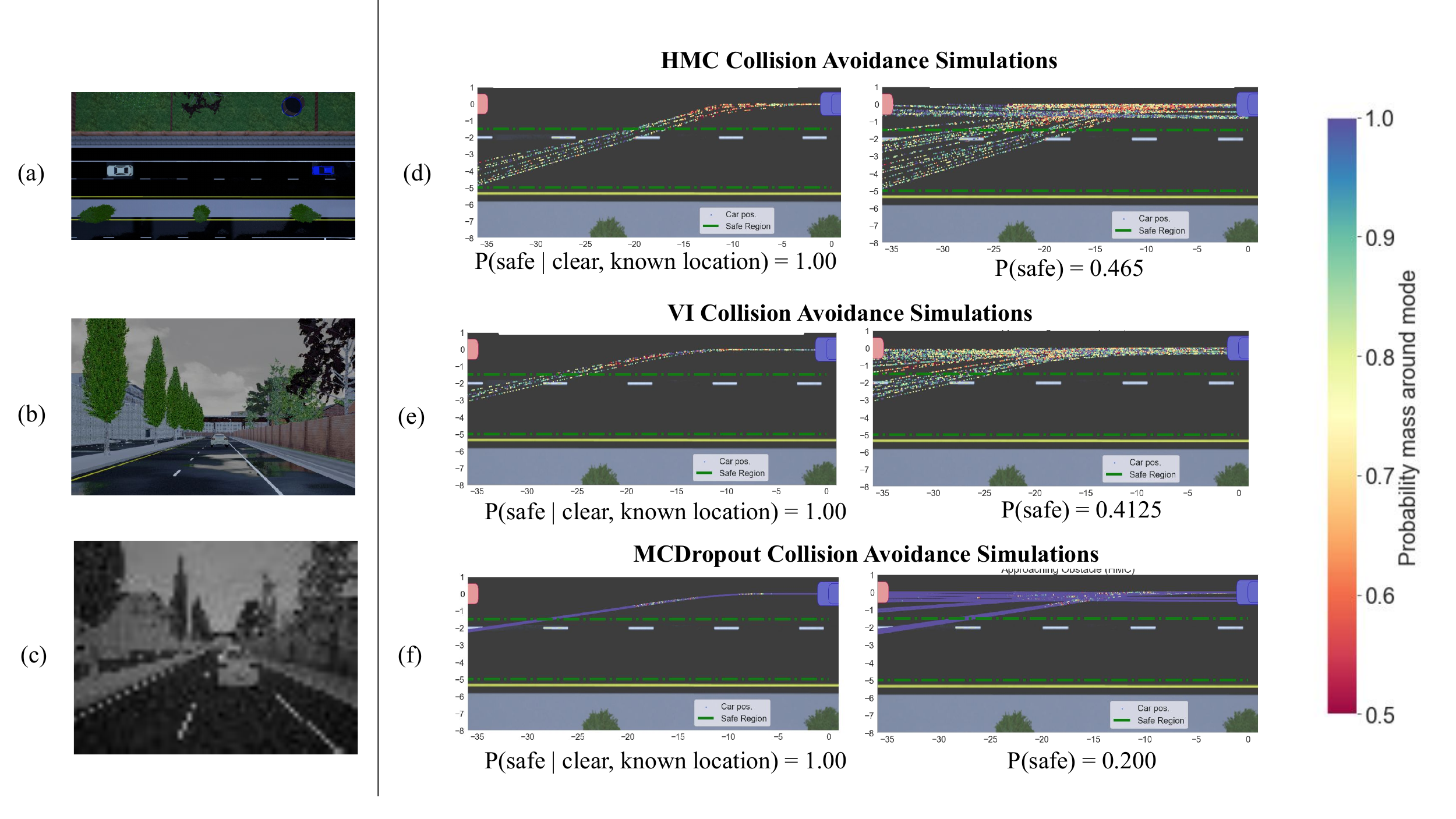}
    \caption{To the left of the black line, we visualize the experimental set up. (a) Spatial distribution of cars; (b) original camera signal; (c) input to BNN. In the centre of the figure we plot the course of the vehicle controlled by a BNN, where each row represents a different posterior approximation (d) HMC, (e) VI, (f) MC Dropout. Each dot, representing the position of the car during its trajectory is colored based on the uncertainty of the controller.}
    \label{fig:coll_avoid_main}
\end{figure*}

In this section, we describe an extendable experimental set up for computing the measures in Definition~\ref{Def:PlanningTImeHorizon} and Definition~\ref{def:ProblemRealTIme}. We first show that use of the measure in Definition~\ref{def:ProblemRealTIme} in conjunction with classical measures of uncertainty can greatly increase the safety of an autonomous vehicle when it is in unfamiliar scenarios. We then consider probabilistic safety as defined in Definition~\ref{Def:PlanningTImeHorizon} and we show that this measure can be effectively used in order to identify problematic scenarios in which further data acquisition should occur.



\subsection{Real-time Collision Avoidance}
\label{sec:realtimeavoid}



In Figure~\ref{fig:coll_avoid_main}, we can see an example of a collision avoidance test set up. We place a vehicle 40 meters away from an obstacle in fixed weather conditions along a single roadway. We then train a BNN controller on data collected from safe human driving in this scenario. Below, we describe a general framework for performing collision avoidance which generalizes to any scenario one would like to test. Further, the system that we use can be implemented for any BNN that is trained to drive autonomously, and can detect situations in which the car is uncertain in order to improve safety.

\begin{figure}
    \centering
    \includegraphics[width=1.0\linewidth]{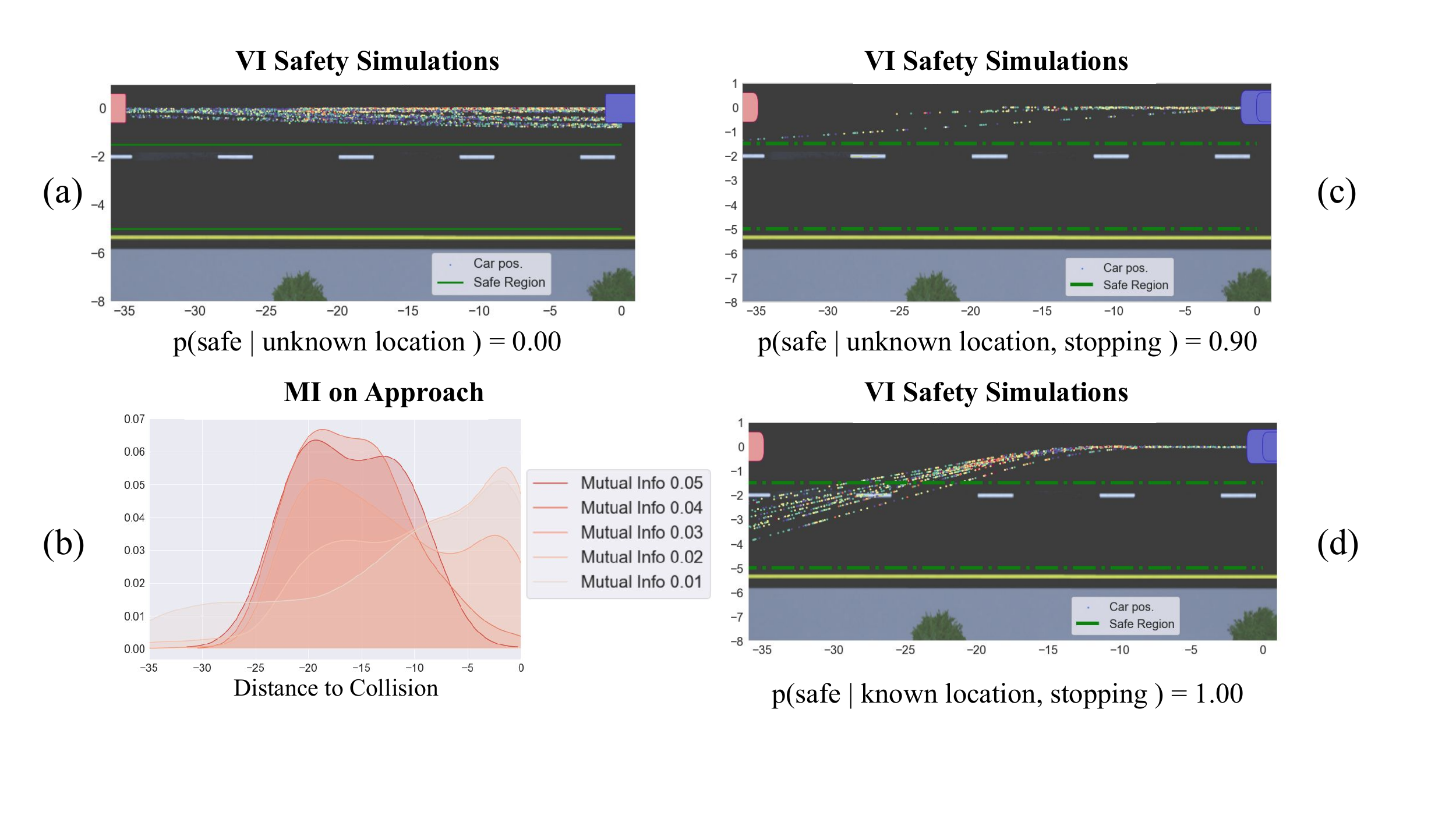}
    \caption{
    Demonstration of how the uncertainty-aware stopping procedure performs. (a) Original safety of VI without stopping algorithm. (b) Mutual information signal spikes as we approach the obstacle. (c) VI safety with stopping. (d) VI performance in a known environment with stopping.}
    \label{fig:warnings}
\end{figure}

The uncertainty-aware decision system is designed in two stages. In the first stage, we simulate more runs of the vehicle driving without any collision avoidance system present. We rely only on the learned behavior of the vehicle (plots of these runs can be seen in Figure~\ref{fig:coll_avoid_main}). At this stage, we are able to qualitatively understand the behavior of each network posterior in terms of the uncertainty it produces as it approaches the obstacle. The behavior of uncertainty can roughly be seen in the bottom left-hand corner of Figure~\ref{fig:warnings}. We note that it is possible, though less desirable, to perform this qualitative evaluation using a held-out, test data set. Because the input we observe at time $t$ depends on all of the decisions made up to that time, 
generating safety or uncertainty estimates based on another controllers decisions may be inaccurate due to the potentially low probability of ever observing those states with the current controller under consideration. In the second stage, we use the captured information about uncertainty in order to generate actionable warning thresholds. For example, if we see that there is typically a large spike in uncertainty as the car approaches the obstacle, we can use a threshold in order to stop the car when we experience a similar peak in the future. 

We use a three tiered warning system based on real-time decision confidence, as defined in Definition \ref{def:ProblemRealTIme}. That is, given an image at time $k$ we bin network decisions into four categories based on the value of $\eta_2$. Often times no warning will be thrown, i.e., $\eta_2\geq \delta_1$ for a given $\delta_1\in [0,1]$. However, in the case that we are less than $\delta_1$-certain ($\eta_2<\delta_1$), a standard warning (warning 1) is thrown. A severe warning (warning 2) is thrown when the network is less that $\delta_2$-certain (this assumes $\delta_2 < \delta_1$). Finally, we consider a warning (warning 0) which is thrown when neither a severe nor standard warning are thrown ($\eta_2\geq \delta_1$), but the predictive distribution exhibits high mutual information, above yet another threshold, in our case 0.45. For our experiments, the constants $\delta_1$ and $\delta_2$ are set to a threshold of 0.7 and 0.6 respectively. The actions that occur at each of these warnings are also configurable. However, we have set up our system such that mutual information warnings slow down the vehicle, standard warnings slow down the vehicle and alert the operator of potential hazard, and severe warnings cause the car to safely brake and alert the operator that they need to assume control of the vehicle. 

Setting these thresholds requires a delicate trade-off between autonomy and safety. If the thresholds are set too low, then the system will operate more autonomously (that is, without asking for user intervention), however it may be less safe. Setting the thresholds too high may be safer, but causes the car to operate less autonomously as the user is constantly prompted for input. In Figure~\ref{fig:warnings}, we show that these sorts of collision avoidance systems can perform well in practice. We show that we can detect and reduce the rate of collision (the inverse of probabilistic safety), improving the safety in unknown conditions from 0.00 ($\pm 0.05$) to  0.90 ($\pm 0.05$), see Figure~\ref{fig:coll_avoid_main}. Moreover, we test that implementation of this strategy does not affect the autonomy of the car in known situations. For this we simulate the situation in which the car was trained and we find that the car still operates with safety probability 1.00, with error margin of 0.05 according to Equation~\ref{eq:chernoff}, and full autonomy (i.e. never stops to ask for user to assume control of the car).

\begin{figure}
    \centering
    \includegraphics[width=0.88\linewidth]{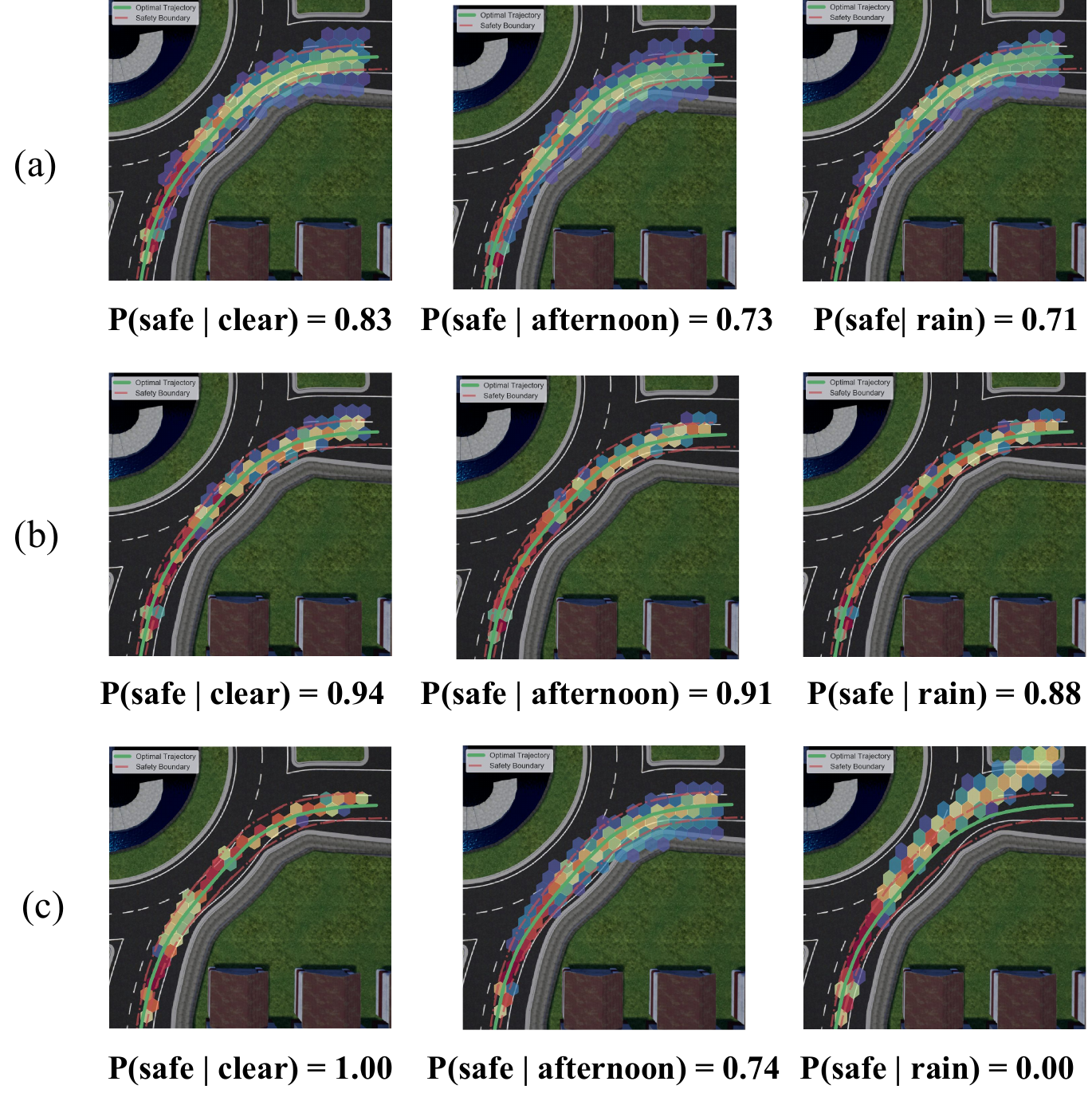}
    \caption{
    Offline safety probabilities of (a) HMC, (b) VI and (c) MCD respectively. Each hexagon is shaded with the probability of the car visiting that area, and the optimal trajectory is plotted with a green line. The red lines show safety boundaries, where outside of this is considered unsafe, and inside is safe.}
    \label{fig:offline_safety_main}
\end{figure}

\subsection{Probabilistic Safety Estimates}

In order to measure the safety of a BNN controller in a particular setting, one must simulate scenarios (e.g. turns, collision avoidance, intersections) in various conditions in order to satisfy the bound in Equation~\ref{eq:chernoff}. Though we do simulations in order to test the safety of a turn, running the correct number of simulations with diverse environmental conditions works on any scenario one would like to test. For example, the notion of probabilistic safety  is also used to calculate the safety in Figure~\ref{fig:coll_avoid_main}.

Figure~\ref{fig:offline_safety_main} shows the test setup for probabilistic safety estimates. We place a vehicle approximately 10 meters from the entrance of a roundabout in fixed weather conditions. We then collect training data using the built in autopilot. The autopilot is set to drive the car through the roundabout, taking the first exit. We then use our safety boundaries to determine the probability that a specific controller will drive safely, that is, stay within our safety boundaries. We are then left with safety probabilities for each section of road tested, for each controller.

While we expect the controller to be able to safely navigate from its trained starting point to the end point in the weather it has seen, we seek to test the robustness of posterior distributions to changes in scenery and weather conditions in order to also include simulations of potential worst-case deployment performance. In row (a) of Figure~\ref{fig:offline_safety_main} we see that, while the variance can be useful in collision avoidance, the wide variance of HMC causes a larger proportion of trajectories to fall outside of the safety boundary. The estimated probability of safety for HMC, across all weathers, was 0.766 ($\pm 0.05$). Row (b) of Figure~\ref{fig:offline_safety_main} reports the consistency of VI across different weather conditions with a cumulative safety probability estimate of 0.91 ($\pm 0.05$) in this particular test case. The main reason for lack of safety in VI was veering into the center lane of the roundabout. Finally, in  row (c) we see the performance of MC Dropout. In the training environment, it was the only method to achieve a perfect safety score; however, we see the network fails to generalize well to other weather conditions. While MC Dropout performs slightly better than HMC in the more dim light of the afternoon, it fails catastrophically in the rain. MCDropout's overall probabilistic safety, prior to the consideration of rain, was 0.87. When we factor rainy environments, the overall probabilistic safety of MCDropout falls to 0.58 ($\pm 0.05$). It is likely that if we were to retrain MC Dropout in all weather conditions and re-run the safety analysis we would see a perfect safety score, as we do currently with clear weather. In this way, we can use our offline safety probability as a guide for active learning in order to increase data coverage and scenario representation in training data.

\section{Conclusion}
We presented a framework for evaluating the safety of end-to-end BNN controllers for self-driving cars, which allows one to obtain uncertainty estimates for the controller's decisions with a priori statistical guarantees.
On experiments performed on the CARLA driving simulator we showed that our statistical framework can be used to evaluate model robustness to changes in weather, location, and observation noise. Further, we illustrate how our results can be successfully employed to detect and avoid a high percentage of collisions.


\bibliographystyle{IEEEtran}
\bibliography{reference}

\begin{thebibliography}{10}
\providecommand{\url}[1]{#1}
\csname url@rmstyle\endcsname
\providecommand{\newblock}{\relax}
\providecommand{\bibinfo}[2]{#2}
\providecommand\BIBentrySTDinterwordspacing{\spaceskip=0pt\relax}
\providecommand\BIBentryALTinterwordstretchfactor{4}
\providecommand\BIBentryALTinterwordspacing{\spaceskip=\fontdimen2\font plus
\BIBentryALTinterwordstretchfactor\fontdimen3\font minus
  \fontdimen4\font\relax}
\providecommand\BIBforeignlanguage[2]{{%
\expandafter\ifx\csname l@#1\endcsname\relax
\typeout{** WARNING: IEEEtran.bst: No hyphenation pattern has been}%
\typeout{** loaded for the language `#1'. Using the pattern for}%
\typeout{** the default language instead.}%
\else
\language=\csname l@#1\endcsname
\fi
#2}}

\bibitem{Waymo}
W.~Team, ``Waymo’s fleet reaches 4 million self-driven miles,''
  \url{https://medium.com/waymo/waymos-fleet-reaches-4-million-self-driven-miles-b28f32de495a},
  2017, accessed: 2018-08-16.

\bibitem{CADis}
C.~DMV, ``Autonomous vehicle disengagement report,''
  \url{https://www.dmv.ca.gov/portal/dmv/detail/vr/autonomous/disengagement_report_2017},
  2017, accessed: 2018-08-16.

\bibitem{TeslaCrash}
D.~T. Danny~Yadron, ``Tesla driver dies in first fatal crash while using
  autopilot mode,''
  \url{https://www.theguardian.com/technology/2016/jun/30/tesla-autopilot-death-self-driving-car-elon-musk},
  2016, accessed: 2018-08-16.

\bibitem{mackay1992practical}
D.~J. MacKay, ``A practical {B}ayesian framework for backpropagation
  networks,'' \emph{Neural computation}, vol.~4, no.~3, pp. 448--472, 1992.

\bibitem{mcallister2017concrete}
R.~McAllister, Y.~Gal, A.~Kendall, M.~Van Der~Wilk, A.~Shah, R.~Cipolla, and
  A.~V. Weller, ``Concrete problems for autonomous vehicle safety: Advantages
  of bayesian deep learning.''\hskip 1em plus 0.5em minus 0.4em\relax
  International Joint Conferences on Artificial Intelligence, Inc., 2017.

\bibitem{bishop2006pattern}
C.~M. Bishop, \emph{Pattern recognition and machine learning}.\hskip 1em plus
  0.5em minus 0.4em\relax springer, 2006.

\bibitem{chernoff1952measure}
H.~Chernoff \emph{et~al.}, ``A measure of asymptotic efficiency for tests of a
  hypothesis based on the sum of observations,'' \emph{The Annals of
  Mathematical Statistics}, vol.~23, no.~4, pp. 493--507, 1952.

\bibitem{Dosovitskiy17}
A.~Dosovitskiy, G.~Ros, F.~Codevilla, A.~Lopez, and V.~Koltun, ``{CARLA}: {An}
  open urban driving simulator,'' in \emph{Proceedings of the 1st Annual
  Conference on Robot Learning}, 2017, pp. 1--16.

\bibitem{bojarski2016end}
M.~Bojarski, D.~Del~Testa, D.~Dworakowski, B.~Firner, B.~Flepp, P.~Goyal, L.~D.
  Jackel, M.~Monfort, U.~Muller, J.~Zhang, \emph{et~al.}, ``End to end learning
  for self-driving cars,'' \emph{arXiv preprint arXiv:1604.07316}, 2016.

\bibitem{gal2016dropout}
Y.~Gal and Z.~Ghahramani, ``Dropout as a bayesian approximation: Representing
  model uncertainty in deep learning,'' in \emph{international conference on
  machine learning}, 2016, pp. 1050--1059.

\bibitem{blundell2015weight}
C.~Blundell, J.~Cornebise, K.~Kavukcuoglu, and D.~Wierstra, ``Weight
  uncertainty in neural networks,'' \emph{arXiv preprint arXiv:1505.05424},
  2015.

\bibitem{neal2011hmc}
R.~M. {Neal}, ``{MCMC using Hamiltonian dynamics},'' \emph{arXiv e-prints}, p.
  arXiv:1206.1901, June 2012.

\bibitem{chen2017end}
Z.~Chen and X.~Huang, ``End-to-end learning for lane keeping of self-driving
  cars,'' in \emph{2017 IEEE Intelligent Vehicles Symposium (IV)}.\hskip 1em
  plus 0.5em minus 0.4em\relax IEEE, 2017, pp. 1856--1860.

\bibitem{bojarski2017explaining}
M.~Bojarski, P.~Yeres, A.~Choromanska, K.~Choromanski, B.~Firner, L.~Jackel,
  and U.~Muller, ``Explaining how a deep neural network trained with end-to-end
  learning steers a car,'' \emph{arXiv preprint arXiv:1704.07911}, 2017.

\bibitem{xu2017end}
H.~Xu, Y.~Gao, F.~Yu, and T.~Darrell, ``End-to-end learning of driving models
  from large-scale video datasets,'' in \emph{Proceedings of the IEEE
  conference on computer vision and pattern recognition}, 2017, pp. 2174--2182.

\bibitem{isermann1984faultdetection}
R.~Isermann, ``Process fault detection based on modeling and estimation
  methods—a survey,'' \emph{automatica}, vol.~20, no.~4, pp. 387--404, 1984.

\bibitem{kendall2017uncertainties}
A.~Kendall and Y.~Gal, ``What uncertainties do we need in bayesian deep
  learning for computer vision?'' in \emph{Advances in neural information
  processing systems}, 2017, pp. 5574--5584.

\bibitem{lee2018ensemble}
K.~Lee, Z.~Wang, B.~I. Vlahov, H.~K. Brar, and E.~A. Theodorou, ``Ensemble
  bayesian decision making with redundant deep perceptual control policies,''
  \emph{arXiv preprint arXiv:1811.12555}, 2018.

\bibitem{kahn2017uncertainty}
G.~Kahn, A.~Villaflor, V.~Pong, P.~Abbeel, and S.~Levine, ``Uncertainty-aware
  reinforcement learning for collision avoidance,'' \emph{arXiv preprint
  arXiv:1702.01182}, 2017.

\bibitem{amini2019variational}
A.~Amini, G.~Rosman, S.~Karaman, and D.~Rus, ``Variational end-to-end
  navigation and localization,'' in \emph{2019 International Conference on
  Robotics and Automation (ICRA)}.\hskip 1em plus 0.5em minus 0.4em\relax IEEE,
  2019, pp. 8958--8964.

\bibitem{huang2019uncertainty}
X.~Huang, S.~McGill, B.~C. Williams, L.~Fletcher, and G.~Rosman,
  ``Uncertainty-aware driver trajectory prediction at urban intersections,''
  \emph{arXiv preprint arXiv:1901.05105}, 2019.

\bibitem{feng2018towards}
D.~Feng, L.~Rosenbaum, and K.~Dietmayer, ``Towards safe autonomous driving:
  Capture uncertainty in the deep neural network for lidar 3d vehicle
  detection,'' in \emph{2018 21st International Conference on Intelligent
  Transportation Systems (ITSC)}.\hskip 1em plus 0.5em minus 0.4em\relax IEEE,
  2018, pp. 3266--3273.

\bibitem{quilbeuf2018statistical}
J.~Quilbeuf, M.~Barbier, L.~Rummelhard, C.~Laugier, A.~Legay, B.~Baudouin,
  T.~Genevois, J.~Iba{\~n}ez-Guzm{\'a}n, and O.~Simonin, ``Statistical model
  checking applied on perception and decision-making systems for autonomous
  driving,'' 2018.

\bibitem{cardelli2019robustness}
L.~Cardelli, M.~Kwiatkowska, L.~Laurenti, and A.~Patane, ``Robustness
  guarantees for bayesian inference with gaussian processes,'' in
  \emph{Proceedings of the AAAI Conference on Artificial Intelligence},
  vol.~33, 2019, pp. 7759--7768.

\bibitem{graves2011practical}
A.~Graves, ``Practical variational inference for neural networks,'' in
  \emph{Advances in neural information processing systems}, 2011, pp.
  2348--2356.

\bibitem{gihman2012controlled}
I.~I. Gihman and A.~V. Skorohod, \emph{Controlled stochastic processes}.\hskip
  1em plus 0.5em minus 0.4em\relax Springer Science \& Business Media, 2012.

\bibitem{chatterjee2016decidable}
K.~Chatterjee, M.~Chmel{\'\i}k, and M.~Tracol, ``What is decidable about
  partially observable markov decision processes with $\omega$-regular
  objectives,'' \emph{Journal of Computer and System Sciences}, vol.~82, no.~5,
  pp. 878--911, 2016.

\bibitem{reynolds2018uber}
\BIBentryALTinterwordspacing
H.~Reynolds, ``Simulation: The invisible gatekeeper,'' \emph{Medium}, 2019.
  [Online]. Available:
  \url{https://medium.com/@UberATG/simulation-the-invisible-gatekeeper-e6ef84ea7647}
\BIBentrySTDinterwordspacing

\bibitem{abate2008probabilistic}
A.~Abate, M.~Prandini, J.~Lygeros, and S.~Sastry, ``Probabilistic reachability
  and safety for controlled discrete time stochastic hybrid systems,''
  \emph{Automatica}, vol.~44, no.~11, pp. 2724--2734, 2008.

\bibitem{Bortolussi:2019:CLM:3347091.3331452}
\BIBentryALTinterwordspacing
L.~Bortolussi, L.~Cardelli, M.~Kwiatkowska, and L.~Laurenti, ``Central limit
  model checking,'' \emph{ACM Trans. Comput. Logic}, vol.~20, no.~4, pp.
  19:1--19:35, July 2019. [Online]. Available:
  \url{http://doi.acm.org/10.1145/3331452}
\BIBentrySTDinterwordspacing

\bibitem{shannon2001mathematical}
C.~E. Shannon, ``A mathematical theory of communication,'' \emph{ACM SIGMOBILE
  mobile computing and communications review}, vol.~5, no.~1, pp. 3--55, 2001.

\bibitem{cardelli2019statistical}
L.~Cardelli, M.~Kwiatkowska, L.~Laurenti, N.~Paoletti, A.~Patane, and
  M.~Wicker, ``Statistical guarantees for the robustness of bayesian neural
  networks,'' \emph{arXiv preprint arXiv:1903.01980}, 2019.

\bibitem{legay2010statistical}
A.~Legay, B.~Delahaye, and S.~Bensalem, ``Statistical model checking: An
  overview,'' in \emph{International conference on runtime verification}.\hskip
  1em plus 0.5em minus 0.4em\relax Springer, 2010, pp. 122--135.

\bibitem{rothe2015dex}
R.~Rothe, R.~Timofte, and L.~Van~Gool, ``Dex: Deep expectation of apparent age
  from a single image,'' in \emph{Proceedings of the IEEE International
  Conference on Computer Vision Workshops}, 2015, pp. 10--15.

\bibitem{ovadia2019can}
Y.~Ovadia, E.~Fertig, J.~Ren, Z.~Nado, D.~Sculley, S.~Nowozin, J.~V. Dillon,
  B.~Lakshminarayanan, and J.~Snoek, ``Can you trust your model's uncertainty?
  evaluating predictive uncertainty under dataset shift,'' \emph{arXiv preprint
  arXiv:1906.02530}, 2019.

\bibitem{gal2017concrete}
Y.~Gal, J.~Hron, and A.~Kendall, ``Concrete dropout,'' in \emph{Advances in
  Neural Information Processing Systems}, 2017, pp. 3581--3590.

\bibitem{tran2016edward}
D.~Tran, A.~Kucukelbir, A.~B. Dieng, M.~Rudolph, D.~Liang, and D.~M. Blei,
  ``Edward: A library for probabilistic modeling, inference, and criticism,''
  \emph{arXiv preprint arXiv:1610.09787}, 2016.

\end{thebibliography}

\end{document}